\documentclass{article} 
\usepackage[preprint]{colm2025_conference}

\usepackage{microtype}
\usepackage{hyperref}
\usepackage{url}
\usepackage{booktabs}

\usepackage{lineno}

\usepackage{tabularx}
\usepackage{booktabs}
\usepackage{pifont}
\usepackage{xcolor}
\usepackage{colortbl}
\usepackage{tcolorbox}
\usepackage{color, colortbl}
\usepackage[normalem]{ulem}

\usepackage{times}
\usepackage{latexsym}
\usepackage{tikz}
\newcommand\encircle[2][]{\tikz[overlay]\node[fill=blue!20,inner sep=2pt, anchor=text, rectangle, rounded corners=1.5mm,#1] {#2};\phantom{#2}}

\usepackage{wrapfig}

\newcommand{\cmark}{\ding{51}} 
\newcommand{\xmark}{\ding{55}}

\definecolor{darkblue}{rgb}{0, 0, 0.5}
\hypersetup{colorlinks=true, citecolor=darkblue, linkcolor=darkblue, urlcolor=darkblue}

\title{Does Learning Mathematical Problem-Solving \\ Generalize to Broader Reasoning?}

\author{Ruochen Zhou$^1$ \quad Minrui Xu$^2$ \quad Shiqi Chen$^1$ \quad Junteng Liu$^2$ \quad Yunqi Li$^2$ \quad Xinxin Lin$^2$  \\
\AND
Zhengyu Chen$^3$ \quad Junxian He$^2$ \\
\\
$^1$City University of Hong Kong  \quad $^2$HKUST  \quad $^3$Meituan
}

\begin{document}

\ifcolmsubmission
\linenumbers
\fi

\maketitle

\begin{abstract}
There has been a growing interest in enhancing the mathematical problem-solving (MPS) capabilities of large language models. 
While the majority of research efforts concentrate on creating specialized models to solve mathematical problems, it remains unknown how learning mathematical problem-solving generalizes to help develop other reasoning abilities. 
In this paper, we present an empirical investigation into the generalization potential of various MPS training approaches, such as continual pretraining, instruction tuning, and rule-based reinforcement learning across various data sources, including both short and long chain-of-thought (CoT) samples.    
Evaluation on 5 mathematical and 8 general reasoning benchmarks show that continual pretraining on math text is able to generalize to general reasoning tasks to some extent. In constrast, instruction tuning on conventional, short MPS samples provides limited benefits and, in many cases, even impairs generalization performance.  
Notably, training with long CoT responses for MPS samples and incorporating rule-based reinforcement learning on MPS queries exhibit distinct behavior, significantly enhancing generalization by extending the model's reasoning processes into other domains. These results suggest that traditional approaches to learning MPS with short reasoning chains largely fail to achieve robust generalization. However, the emerging paradigm of longer reasoning chains, coupled with self-reflection, offers a promising direction for improving generalized reasoning abilities through learning from specialized domains.

\end{abstract}

\section{Introduction}

\begin{wrapfigure}{r}{0.48\textwidth}
    \centering
    \vspace{-10mm} 
    \includegraphics[width=\linewidth]{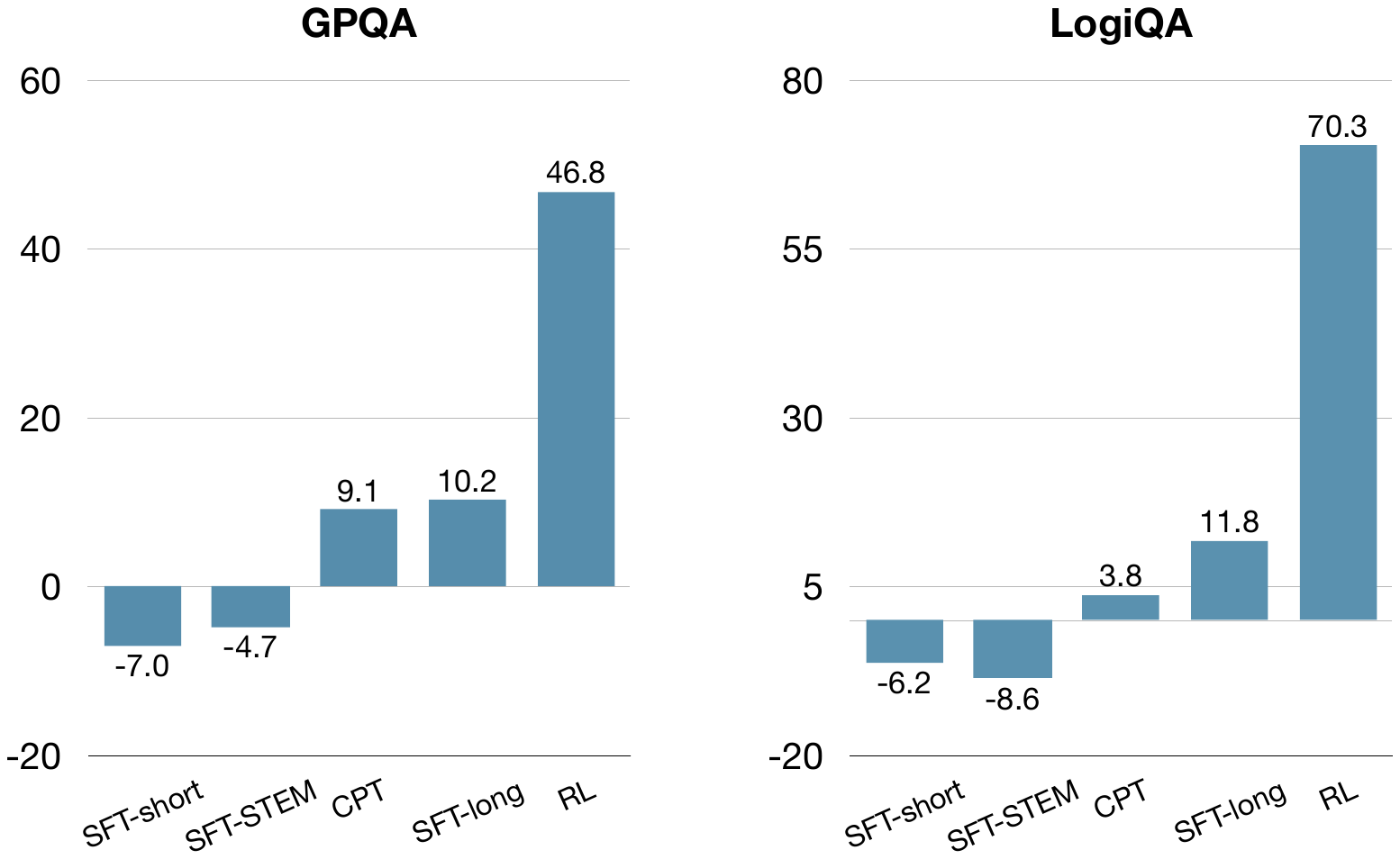}
    
    \vspace{-4mm} 
    \caption{
        Relative gain (\%) on GPQA and LogiQA of five different MPS training approaches. SFT-short: Supervised fine-tuning on short CoT data. SFT-STEM: Supervised fine-tuning on STEM-related data. CPT: Continual pretraining. SFT-long: Supervised fine-tuning on long CoT data. RL: Rule-based reinforcement learning on MPS queries.
    }
    \vspace{-4mm} 
    \label{fig:fig_relative_logic}
\end{wrapfigure}
Cognitive neuroscience research has demonstrated that learning to solve mathematical problems enhances general reasoning abilities in humans by promoting logical thinking, abstract reasoning, and transferable problem-solving strategies across various domains~\citep{dehaene2004arithmetic,hawes2020explains}. 

This notion -- that learning math fosters the development of general reasoning skills 
-- points toward a ``\emph{math for AI}'' vision,
where incorporating mathematical reasoning data into AI training could help large language models (LLMs) develop more complex and versatile reasoning abilities. The ``math for AI'' goal is particularly relevant to recent attentions to complex reasoning abilities of LLMs~\citep{openaio1,deepseek-r1}, as mathematical problem-solving (MPS) is one of the few domains where large volumes of long and intricate CoT data can be 
generated and verified~\citep{tang2024mathscale,lu2024mathgenie}, making it a valuable data source to potentially learn complex reasoning.
However, while numerous models have been developed to tackle mathematical problem-solving
~\citep{luo2023wizardmath,numina_math_datasets,ye2025limo}
, their evaluations focus narrowly on MPS benchmarks, it is unclear whether these approaches and the accompanied datasets can really help learn other types of reasoning.
Thus, a key question remains: \emph{Does learning mathematical problem-solving contribute to the development of a model's general reasoning abilities, or does it merely enhance performance on MPS benchmarks?}

\begin{figure*}[!t]
  \centering
  \vspace{-16mm}
  \includegraphics[width=0.95\linewidth]{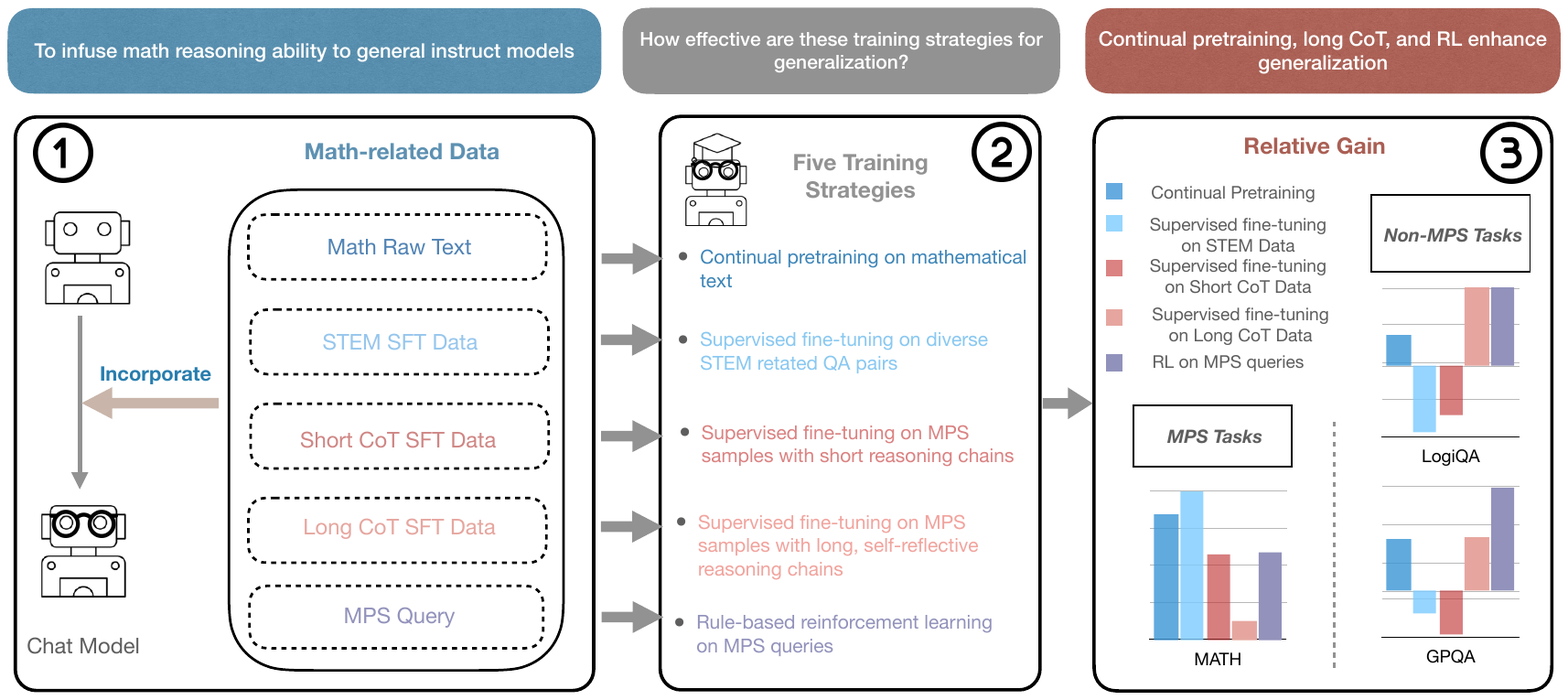}
    \vspace{-16mm}
  \caption{An overview of incorporating mathematical reasoning ability into general instruction models. ~\encircle[fill=white, text=black, draw=black, line width=1.5pt]{ 1}: Math-related data, including raw text, STEM SFT datasets, short CoT SFT datasets, long CoT SFT datasets and MPS queries, are incorporated into a chat model. ~\encircle[fill=white, text=black, draw=black, line width=1.5pt]{ 2}: Different training strategies based on the five kinds of math-related data are explored for generalization. ~\encircle[fill=white, text=black, draw=black, line width=1.5pt]{ 3}: Evaluation results observed in both MPS and non-MPS tasks.}
  \label{fig:fig_main}
\vspace{-2mm}
\end{figure*}

In this study, we conduct empirical analysis focusing on this central question as shown in Figure~\ref{fig:fig_main}. Specifically, we explore whether training LLMs on mathematical problem-solving tasks can help broader reasoning tasks beyond mathematics. We focus on five common training strategies to enhance LLMs' capabilities in solving mathematical problems
:  
(1) \emph{Continual pretraining on mathematical text} involves extending the pretraining of LLMs on large-scale mathematical text to enhance their adaptability to the mathematical domain~\citep{lin2024rho1,deepseek-math};
(2) \emph{Supervised fine-tuning (SFT) on STEM-related data} involves training models on diverse question-answer pairs collected from a wide range of STEM disciplines which help improve the model’s general reasoning skills~\citep{yue2024mammoth2scalinginstructionsweb, cheng2024instruction}.
(3) \emph{Supervised fine-tuning (SFT) on MPS samples with short reasoning chains} focuses on directly training models on mathematical problem-solving (MPS) datasets where solutions are presented in concise, step-by-step forms. This approach is widely used and has been shown to achieve strong baseline performance across many MPS benchmarks~\citep{yu2023metamath,numina_math_datasets,tong2024dartmathdifficultyawarerejectiontuning,ye2025limo}. However, its limitation lies in the lack of deeper reflection or explanation within the reasoning process.
(4) \emph{Supervised fine-tuning (SFT) on MPS samples with long, self-reflective reasoning chains} represents a recent paradigm shift toward enhancing models' abilities to produce extended and introspective reasoning.  These chains often include detailed verification steps and self-checking behaviors
~\citep{openaio1,deepseek-r1}. Existing works have demonstrated that the MPS abilities can be greatly boosted with around 1000 examples~\citep{ye2025limo,s11}. 
and (5) \emph{Rule-based reinforcement learning on MPS queries} introduces a rule-based reward mechanisms to improve models' reasoning ability. Recent works have shown that RL-based methods can facilitate the emergence of long reasoning chains and yield substantial improvements in mathematical problem-solving~\citep{deepseek-r1,zeng2025simplerlzooinvestigatingtamingzero}.

We perform control experiments and evaluate a series of models created by the five training strategies above, where the models are either from open-source checkpoints or our own training. 
We assess these models across multiple benchmarks involving 
both math-related and general reasoning tasks.
Our experimental results reveal that conventional supervised fine-tuning on short reasoning chains generalizes poorly and often hurts the performance on non-mathematical reasoning tasks, while continual pretraining on raw mathematical text improves performance on 6 out of 8 general reasoning tasks. 
However, supervised fine-tuning using long, self-reflective reasoning chains exhibits markedly different behavior. For instance, models like LIMO~\citep{ye2025limo} consistently enhance the performance of the base model, Qwen2.5-32B-Instruct~\citep{qwen2.5}, on general reasoning tasks, even when trained on as few as 817 MPS examples. The gains are particularly pronounced on certain benchmarks. As shown in Figure~\ref{fig:fig_relative_logic}, LIMO achieves a relative improvement of 10.2\% on GPQA~\citep{rein2023gpqa} and 11.8\% on LogiQA~\citep{liu2020logiqa}. Upon careful analysis, we find that this kind of tuning with long CoT reasoning chains on MPS examples significantly increases the response length across other domains. This activation of the model's ``long reasoning mode'' appears to enable better generalization across diverse reasoning tasks. Furthermore, RL with rule-based constraints on MPS queries also exhibits improvement on both mathematical and general reasoning tasks. Specifically, SimpleRL-Zero and SimpleRL~\citep{zeng2025simplerl}, consistently outperform both the base model (Qwen2.5-Math-7B) and the Math Data SFT model across multiple tasks.

On a separate line, we further conduct a preliminary study to explore other training tasks that might generalize better than mathematical problem-solving. Specifically, we evaluate several public, reasoning-intensive supervised fine-tuning datasets such as \emph{Magicoder-Evol-Instruct}~\citep{wei2023magicoder}, \emph{Magpie-Reasoning}~\citep{xu2024magpie}, and \emph{OpenOrca}~\citep{OpenOrca}. 
Unlike mathematical problem-solving, these datasets typically span a broader range of reasoning domains, such as code generation and logical reasoning. However, none of them succeed in achieving satisfactory generalization across a wide array of tasks, indicating that new training objectives may be necessary to substantially improve generalization beyond what is achievable through mathematical problem-solving.

We encourage future research to rethink the objectives when studying mathematical reasoning. If the goal is to enhance general reasoning capabilities -- rather than obtaining a math-specialized model -- it is important to consider how learning mathematical problem solving can generalize.

\section{Common Approaches for Learning MPS and Model Setup}
\label{sec:train}

In this study, we examine five common approaches for learning mathematical problem-solving, mainly involving continual pretraining, several supervised fine-tuning or distillation variants and RL with rule-based rewards across different data formats.
Due to the expensive cost of running some of the training paradigms, we obtain the required model from either the open source checkpoints or our own training.

\paragraph{Continual Pretraining on Mathematical Text.}  
In mathematics, where texts often involve multi-step reasoning and formal expressions, this approach helps models better grasp the reasoning patterns~\citep{lewkowycz2022solving}. 
Due to the expensive cost of running continual pretraining, in this study, we experiment with two open-weight LLMs continually pretrained on mathematical-related text: RhO-Math~\citep{lin2024rho1} and DeepSeekMath-Base~\citep{deepseek-math}. 
DeepSeekMath-Base is continually pretrained based on the DeepSeek-Coder-Base~\citep{guo2024deepseek} model using a large mathematical corpus called the DeepSeekMath Corpus. 
Rho-Math-7B is from continual pretraining with Selective Language Modeling method through the OpenWebMath corpus~\citep{paster2023openwebmath} in Mistral-7B~\citep{jiang2023mistral7b}.
Distinct from normal continual pretraining, Rho-Math utilizes another reference model to select tokens and only optimizes losses on the selected tokens.

\paragraph{Supervised Fine-tuning (SFT) on STEM-related Data.} 

Supervised fine-tuning (SFT) on STEM-related data improves a model’s generalization across diverse tasks while enhancing its instruction-following capabilities~\citep{yue2024mammoth2scalinginstructionsweb,chung2024scaling,cheng2024instruction}. This approach involves with large STEM-related QA datasets, often synthesized from raw text, encompassing various formats, complexities, and problem types. In our study, we leverage the open-weight MammoTH2 model~\citep{yue2024mammoth2scalinginstructionsweb} to evaluate it on broader tasks. MammoTH2 was trained on approximately 10 million QA pairs synthesized through open-source LLMs from a wide range of mathematical, science and engineering texts.

\paragraph{Supervised Fine-tuning (SFT) on MPS samples with short reasoning chains.} 
Unlike continual pretraining or instruction pretraining on diverse math-related QA pairs, this approach focuses on smaller, domain-specific datasets typically aligned with benchmark tasks.
This is the most commonly used approach to boost MPS scores due to its efficiency. 
To assess whether models finetuned on MPS datasets can generalize beyond their source tasks, we first use two different normal MPS-oriented datasets to train two models on our own: a chain-of-thought (CoT,~\citet{wei2022chain}) model Math-CoT SFT, and a program-of-thought (PoT,~\citet{chen2022program}) Math-PoT SFT.
Math-CoT SFT was trained on the MetaMath dataset~\citep{yu2023metamath}, which draws primarily from the GSM8K and MATH benchmarks, all structured in a chain-of-thought format. Math-PoT SFT, on the other hand, was trained on the NuminaMath-TIR dataset~\citep{numina_math_datasets}, which includes problems from GSM8K and MATH, as well as other mathematical benchmarks, with tasks presented in natural language and solutions in code snippets.
The NuminaMath-TIR dataset directly leads to the NuminaMath model that wins a recent AI for Math competition.\footnote{\href{https://www.kaggle.com/competitions/ai-mathematical-olympiad-prize/leaderboard}{https://www.kaggle.com/competitions/ai-mathematical-olympiad-prize/leaderboard}}

\paragraph{Supervised Fine-tuning (SFT) on MPS samples with long, self-reflective reasoning chains.}
There has been a growing interest in developing models with long reasoning abilities since the release of o1 models~\citep{openaio1}.
These long-reasoning models are achieving state-of-the-art performance on most MPS benchmarks, greatly outperforming their short-reasoning counterparts~\citep{deepseek-r1,s11,ye2025limo}. 
Different from conventional, short reasoning chains, the long reasoning chains often consist of self-reflection processes where the models validate and try to correct their own responses. This is believed to be highly helpful to solve challenging mathematical questions. 
Related to this study,~\citet{huang2024o1replicationjourney} find that distilling long reasoning chains on mathematical problem-solving increases the response length and arouses the self-reflective behaviors in other domains as well.
In this work, we study instruction tuning with long reasoning chains of MPS examples with two representative models, s1~\citep{s11} and LIMO~\citep{ye2025limo}. They are both fine-tuned based on Qwen2.5-32B-Instruct with 1000 and 817 training samples respectively. 
The training data for LIMO consists exclusively of MPS examples, whereas the training data for s1 is more diverse, comprising primarily MPS examples along with a small fraction of standardized test data like SAT covering topics such as English, law, and logic.

\paragraph{Reinforcement Learning with MPS samples.}
Beyond instruction tuning, reinforcement learning has been demonstrated as an effective approach to enhancing a model’s reasoning abilities~\citep{deepseek-r1,kimi-1.5}. Unlike conventional reward model-based reinforcement learning, this method often relies on rule-based rewards to guide the learning process.
In this study, we adopt the SimpleRL-Zero and SimpleRL models~\citep{zeng2025simplerl}, which incorporate key rule-based constraints to improve reasoning quality. One crucial rule mandates that the final answer must be enclosed within the correct answer format like \verb|\boxed{}|; failure to comply results in a penalty. Additionally, the correctness of the enclosed answer is verified against the ground truth, granting a positive reward if correct. This kind of reinforcement learning strategy enhances the reliability and precision of mathematical problem-solving in MPS tasks.

\section{Experiments}
In this section, we assess the generalization capabilities across multiple types of reasoning benchmarks of these models, encompassing MPS, math-related (excluding problem-solving) and general reasoning tasks.

\setlength{\tabcolsep}{2pt}
\renewcommand{\arraystretch}{1.1}

\begin{table*}[!t]
\centering
\resizebox{1.0\linewidth}{!}{
\begin{tabular}{l|cc@{\hspace{10pt}}|@{\hspace{5pt}}ccccc}
\toprule 
& \multicolumn{2}{c|@{\hspace{5pt}}}{Models Based on DeepSeek-Coder} & \multicolumn{5}{c}{Models Based on Mistral-7B} \\
\cmidrule(lr){2-3} \cmidrule(lr){4-8}
\textbf{Benchmarks} & \textbf{DeepSeek-Coder} & \textbf{DeepSeekMath} & \textbf{Mistral-7B}   & \textbf{Rho-Math-7B} & \textbf{MAmmoTH2-7B} & \textbf{Math-CoT SFT} & \textbf{Math-PoT SFT} \\
\midrule
\midrule
\multicolumn{8}{c}{\textit{Math Problem-solving Tasks}} \\
\midrule
GSM8K &48.8 &\cellcolor[HTML]{84C96F}66.7 &40.7  &\cellcolor[HTML]{84C96F}64.8 &\cellcolor[HTML]{84C96F}63.6 &\cellcolor[HTML]{84C96F}72.4 &\cellcolor[HTML]{84C96F}67.9 \\
MATH &18.7 &\cellcolor[HTML]{84C96F}34.8 &12.3  &\cellcolor[HTML]{84C96F}29.6 &\cellcolor[HTML]{84C96F}32.9 &\cellcolor[HTML]{84C96F}22.5 &\cellcolor[HTML]{84C96F}28.3 \\
GSM8K MQA &55.1 &\cellcolor[HTML]{84C96F}70.1 &56.7  &\cellcolor[HTML]{84C96F}65.7 &\cellcolor[HTML]{84C96F}72.7 &\cellcolor[HTML]{84C96F}74.8 &\cellcolor[HTML]{84C96F}65.8 \\

\midrule
\multicolumn{8}{c}{\textit{Math-Related Tasks (Excluding Problem-solving)}} \\
\midrule

DocMath &15.0 &\cellcolor[HTML]{ACD79E}18.5 &11.3  &\cellcolor[HTML]{D3EBCB}11.7 &\cellcolor[HTML]{84C96F}19.7 &\cellcolor[HTML]{D3EBCB}13 &\cellcolor[HTML]{D3EBCB}13.5 \\

MR-BEN-math &35.2 &\cellcolor[HTML]{FADADA}34.3 &21.5  &\cellcolor[HTML]{ACD79E}26.8 &\cellcolor[HTML]{ACD79E}23.0 &\cellcolor[HTML]{FADADA}21.2 &\cellcolor[HTML]{FADADA}21.2 \\
\midrule
\multicolumn{8}{c}{\textit{General Reasoning Tasks}} \\
\midrule
ZebraLogic &4.7 &\cellcolor[HTML]{D3EBCB}5.1 &4.8  &\cellcolor[HTML]{ACD79E}6.1 &4.8 &\cellcolor[HTML]{ACD79E}6.3 &\cellcolor[HTML]{ACD79E}7.3 \\
LogiQA &26.3 &\cellcolor[HTML]{ACD79E}27.3 &37.0  &\cellcolor[HTML]{dff8d5}37.2 &\cellcolor[HTML]{F19A9A}33.8 &\cellcolor[HTML]{F19A9A}34.9 &\cellcolor[HTML]{F19A9A}36.7 \\
ProofWriter &34.8 &\cellcolor[HTML]{F4B6B6}32.2 &32.5 &\cellcolor[HTML]{FADADA}32.0 &\cellcolor[HTML]{ACD79E}35.5 &\cellcolor[HTML]{D3EBCB}32.8 &\cellcolor[HTML]{D3EBCB}34.5 \\
GPQA &23.2 &\cellcolor[HTML]{dff8d5}25.3 &30.0  &\cellcolor[HTML]{F19A9A}25.3 &\cellcolor[HTML]{F4B6B6}28.6 &\cellcolor[HTML]{F19A9A}27.9 &\cellcolor[HTML]{F4B6B6}28.7 \\
MMLU-STEM &47.5 &\cellcolor[HTML]{ACD79E}54.3 &54.0  &\cellcolor[HTML]{dff8d5}54.3 &\cellcolor[HTML]{ACD79E}56.7 &\cellcolor[HTML]{F4B6B6}52.6 &\cellcolor[HTML]{FADADA}53.5 \\

WinoGrande &64.5 &\cellcolor[HTML]{FADADA}63.5 &75.8  &\cellcolor[HTML]{F19A9A}71.0 &\cellcolor[HTML]{E17272}70.3 &\cellcolor[HTML]{F19A9A}72.9 &\cellcolor[HTML]{E17272}70.9 \\
ARC-c &40.1 &\cellcolor[HTML]{84C96F}46.1 &54.1  &\cellcolor[HTML]{F19A9A}50.0 &\cellcolor[HTML]{ACD79E}56.5 &\cellcolor[HTML]{F19A9A}52.7 &\cellcolor[HTML]{F19A9A}53.4 \\
BBH &56.8 &\cellcolor[HTML]{ACD79E}60.6 &55.3  &\cellcolor[HTML]{ACD79E}57.0 &\cellcolor[HTML]{ACD79E}56.4 &\cellcolor[HTML]{F19A9A}53.8 &\cellcolor[HTML]{F19A9A}54.1 \\

\bottomrule
\end{tabular}}
\caption{
Performance of different models. The baseline models (DeepSeek-Coder and Mistral-7B) are trained with the general conversation data. The Math-CoT SFT and Math-PoT SFT are trained with the mixed data which combined the math data and the general conversation data. The baseline of DeepSeekMath is DeepSeek-Coder. The baseline of other models is Mistral-7B. \textcolor[HTML]{84C96F}{Green} cells indicate improvement, and \textcolor[HTML]{E17272}{red} cells indicate a decrease.}

\label{table:main_result}
\end{table*}

\subsection{General Setup}
\label{sec:train_setup}
\paragraph{Mimicking the Realistic Setting.}
If the goal is to develop general-purpose models, in practice, the developers typically need to adopt general SFT datasets to align the models with users in the last stage. 
Since our objective is to study whether mixing MPS training during the model development would help promote general reasoning abilities, a general SFT dataset is incorporated in most of our evaluations settings to mimic the realistic setting. 
This setup is often necessary as well as we find that fine-tuning models with hundreds of thousands of MPS examples in the last stage could easily overfit the models on MPS tasks, where the models cannot understand other non-MPS instructions anymore. 
Specifically, for continual pretraining and supervised fine-tuning on STEM data, as these two approaches typically serve as an intermediate stage to obtain an enhanced base model followed by general SFT training~\citep{deepseek-math,yue2024mammoth2scalinginstructionsweb}, we perform instruction tuning on a general conversation data, UltraChat~\citep{ding2023enhancing}, based on DeepSeekMath-Base, RhO-Math, and MammoTH2 models to align and then evaluate them.
For instruction tuning on MPS samples with short reasoning chains, we mix the MPS data with the same amount of UltraChat examples for training, a common practice to compose multiple SFT sources.
For s1 and LIMO models that represent long-reasoning instruction tuning, however, we do not further incorporate general SFT because (1) their base models are Qwen2.5-32B-Instruct that already experiences large-scale general alignment; and (2) their MPS training datasets are small with no more than 1000 examples, where the overfitting issue is not prominent. 

\paragraph{Baselines.}
Our goal is to study whether incorporating these MPS learning components will help improve general reasoning abilities, or only lift performance on MPS benchmarks. Thus for each of the studied approach, the baseline is their respective counterpart without the dedicated math training part. 
For example, for the ``DeepSeekMath-base $\rightarrow$ UltraChat SFT'' model to study the effect of continual pretraining on math text, the baseline is ``DeepSeekCoder-base $\rightarrow$ UltraChat SFT'' where DeepSeekCoder-base is the base model of DeepSeekMath-base before continual pretraining; for the mixed ``MPS+UltraChat SFT'' model, the corresponding baseline is the same base model fine-tuned on UltraChat only.

\paragraph{Data and Models.}
 As briefly introduced in \S\ref{sec:train}, we detail the models that we will evaluate: (1) For continual pretraining on mathematical text, we leveraged two existing checkpoints: \texttt{deepseek-math-7b-base} and \texttt{rho-math-7b-v0.1}. Their corresponding base models, are Deepseek-Coder-Base and Mistral-7B, respectively. (2) For supervised fine-tuning on STEM-related data, we used the checkpoint \texttt{MAmmoTH2-7B}, and Mistral-7B serves as its base model. (3) For supervised fine-tuning on MPS data with short reasoning chains, we fine-tuned the base model \texttt{mistral-7b-v0.1} ourselves using the MetaMath~\citep{yu2023metamath} and NuminaMath-TIR~\citep{numina_math_datasets} datasets mixed with UltraChat to get the Math-CoT SFT model and the Math-PoT SFT model.
 (4) For supervised fine-tuning on MPS data with long reasoning chains, we directly use the s1.1 and LIMO checkpoints, which are both fine-tuned from Qwen2.5-32B-Instruct. 
 We set the model’s maximum generation length to 8192 tokens for both s1.1 and LIMO, ensuring that the models have sufficient context to fully generate their reasoning steps.
 The UltraChat data used in our experiments consists of approximately 200K samples. See more training details in Appendix~\ref{appendix:training_hyperparameters}.

 \paragraph{Evaluation Datasets.}
To assess the models' reasoning capabilities across multiple dimensions, we select three categories of tasks for evaluation: mathematical problem-solving (MPS), math-related reasoning (excluding problem-solving), and general reasoning tasks. The general reasoning tasks cover a wide range of reasoning domains, such as logical reasoning, STEM reasoning, commonsense reasoning, and agent reasoning. Specific benchmarks used include GSM8K~\citep{cobbe2021gsm8k}, MATH~\citep{hendrycksmath2021} and GSM8K MQA for MPS tasks. The GSM8K MQA dataset is a repurposed version of the original GSM8K format, converted into a multiple-choice question format. For math-related reasoning (excluding problem-solving), we used DocMath~\citep{zhao2024docmatheval} and MR-BEN-math. The MR-BEN-math focuses exclusively on the math subject from the MR-BEN~\citep{zeng2024mrben} dataset. In the general reasoning category, we used benchmarks such as ZebraLogic~\citep{zebralogic2024}, LogiQA~\citep{liu2020logiqa}, ProofWriter~\citep{tafjord2020proofwriter}, GPQA~\citep{rein2023gpqa}, MMLU-STEM~\citep{hendryckstest2021}, WinoGrande~\citep{sakaguchi2021winogrande}, ARC-challenge~\citep{allenai:arc}, and BBH~\citep{suzgun2022challenging}. More details about the benchmarks can be found in Appendix~\ref{appendix:benchmarks}.

\subsection{Results of Conventional Approaches}
\label{sec:results}

\begin{figure*}[!t]
    \centering
    \includegraphics[width=1\linewidth]{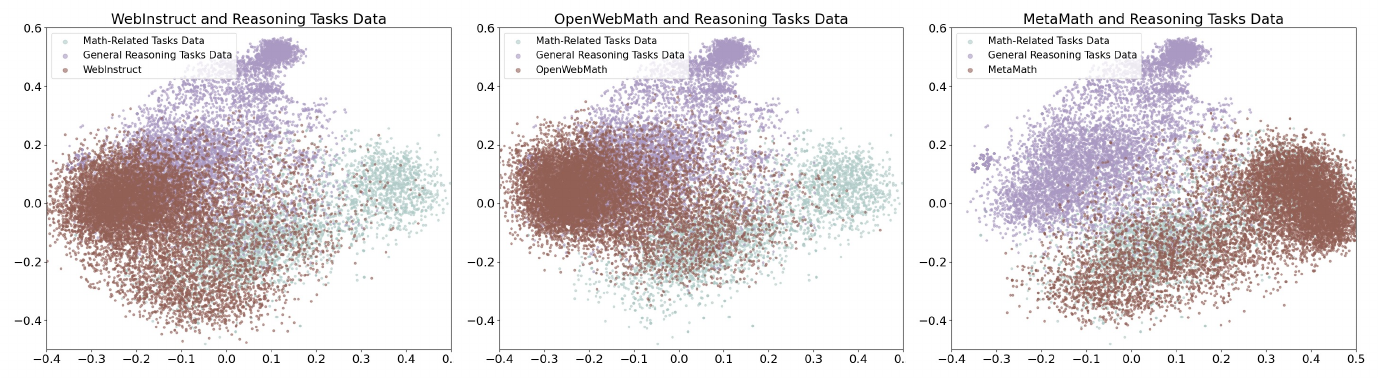}
    \caption{Visualization of the query embedding distributions: WebInstruct, MetaMath, and OpenWebMath compared with all reasoning benchmark datasets. The training data is represented in \textcolor[HTML]{936056}{dark red}, math-related tasks in \textcolor[HTML]{b0ccc8}{green}, and general reasoning tasks in \textcolor[HTML]{aa99c3}{purple}.Embeddings are projected into a 2D space using PCA.}
    \label{fig:fig_dis}
\end{figure*}

In this section, we first report the three relatively conventional approaches without involving long reasoning chains. 
Table~\ref{table:main_result} presents the performance of models of three types of training strategies on different kinds of reasoning tasks.
\paragraph{Learning mathematical problem-solving helps mathematical reasoning in general.}
Unsurprisingly, we observe that all models demonstrate improvements on math-related tasks, including MPS and those not focused on math problem-solving. While the gains on excluding problem-solving math tasks are not as significant as those on MPS tasks, there is still a noticeable improvement. This indicates that learning mathematical problem-solving is able to help improve the model's generalization abilities within the math domain. However, it is worth noting that the two instruction-tuned on MPS data models exhibit a slight decrease in performance on the GSM8K MQA dataset, which implies that these models may struggle with adapting to different question formats This suggests a potential limitation in the ability of instruction tuning to generalize across various task formats, even within the math domain.

\paragraph{Models fine-tuning only on MPS data are less effective at enhancing the model’s general reasoning abilities compared to models training on diverse data.} 
Although no training strategy shows improvements across all benchmarks, we can observe that models trained on diverse data(continual pretraining and fine-tuning on STEM data) generally perform better on a larger number of tasks. For instance, DeepSeekMath demonstrates improvements in six general reasoning tasks. In contrast, models trained with math SFT data experience performance declines in nearly all general reasoning tasks. Specifically, Math-CoT SFT declines in seven tasks, while Math-PoT SFT declines in eight tasks. This suggests that, unlike continual pretraining, instruction tuning only on MPS does not support the generalization of reasoning capabilities beyond mathematical problem-solving tasks.

\paragraph{Greater coverage for pretraining dataset than instruction tuning dataset.} 
To intuitively understand why certain data such as OpenWebMath with DeepSeekMath and WebInstruct with MammoTH2 outperform other MPS SFT datasets on general reasoning. We perform a data coverage analysis to visualize the overlapping relationship between the training data and benchmarks.
Specifically, we sample 1,000 queries from each benchmark and 10,000 queries from the training dataset. 
For math-related training datasets, \emph{WebInstruct}~\citep{yue2024mammoth2scalinginstructionsweb} was used for supervised fine-tuning on STEM data, \emph{MetaMath} was used for SFT on MPS data, and \emph{OpenWebMath}~\citep{paster2023openwebmath} was used for continual pretraining. 
To visualize the data in a 2-dimensional space, we applied Principal Component Analysis (PCA) to reduce the dimensionality of the embeddings. As shown in Figure~\ref{fig:fig_dis}, the query distribution from WebInstruct and OpenWebMath demonstrates more overlap with general reasoning tasks, suggesting that WebInstruct and OpenWebMath encompass a wider variety of topics or problem types that align well with the benchmarks. This overlap likely enhances its effectiveness in generalization tasks. In contrast, MetaMath queries are more concentrated within math-related areas, which may limit its potential for generalization.

\setlength{\tabcolsep}{3pt}
\begin{table*}[!t]
\centering
\resizebox{1.0\linewidth}{!}{
\begin{tabular}{l|cc|cccccc}
\toprule 
& \multicolumn{2}{c|}{MPS Tasks} & \multicolumn{6}{c}{General Reasoning Tasks} \\
\cmidrule(lr){2-3} \cmidrule(lr){4-9}
\textbf{Models}  & \textbf{GSM8K}  & \textbf{MATH} & \textbf{LogiQA} & \textbf{GPQA\_diamond} & \textbf{ARC-c} & \textbf{ZebraLogic} & \textbf{CRUX} & \textbf{HumanEval} \\
\midrule

Qwen2.5-32B-Instruct & \textbf{95.9}(252.1) & 82.7(495.6) & 65.4(413.8) & 50.0(442.8) & 93.3(254.3) & 26.4(683.0) & 69.3(166.8) & 87.2(172.7) \\
\midrule
\multicolumn{9}{c}{\textit{Supervised fine-tuning on MPS samples with long CoT}} \\
\midrule
LIMO & 94.5(1165.4) & 89.3(2011.9) & 73.1(2302.9) & 55.1(3989.5) & \textbf{95.9}(891.8) & \textbf{44.5}(2709.6) & \textbf{77.5}(2389.5) & \textbf{88.4}(228.3) \\
s1.1 & 94.9(1676.6) & \textbf{90.2}(2941.7) & \textbf{73.9}(3854.1) & \textbf{60.6}(4318.3) & 95.4(1855.9) & 37.0(2970.6) & 65.9(2359.2) & 87.8(295.2) \\
\midrule
\multicolumn{9}{c}{\textit{Supervised fine-tuning on MPS samples with short CoT}} \\
\midrule
s1.1-short & 93.4(289.9) & 81.3(391.5) & 59.6(207.8) & 53.5(311.6) & 80.6(171.5) & 21.1(528.6) & 52.1(307.0) & 85.4(225.8) \\
\bottomrule
\end{tabular}}
\caption{
Accuracy (tokens in response) of Qwen2.5-32B-Instruct, LIMO, s1.1 and s1.1-short
}
\vspace{-2mm}
\label{table:qwen_result}
\end{table*}

\subsection{Results of Long-CoT Training}
\label{sec:long-cot}

\begin{figure*}[t]
  \centering
  \includegraphics[width=1.0\linewidth]{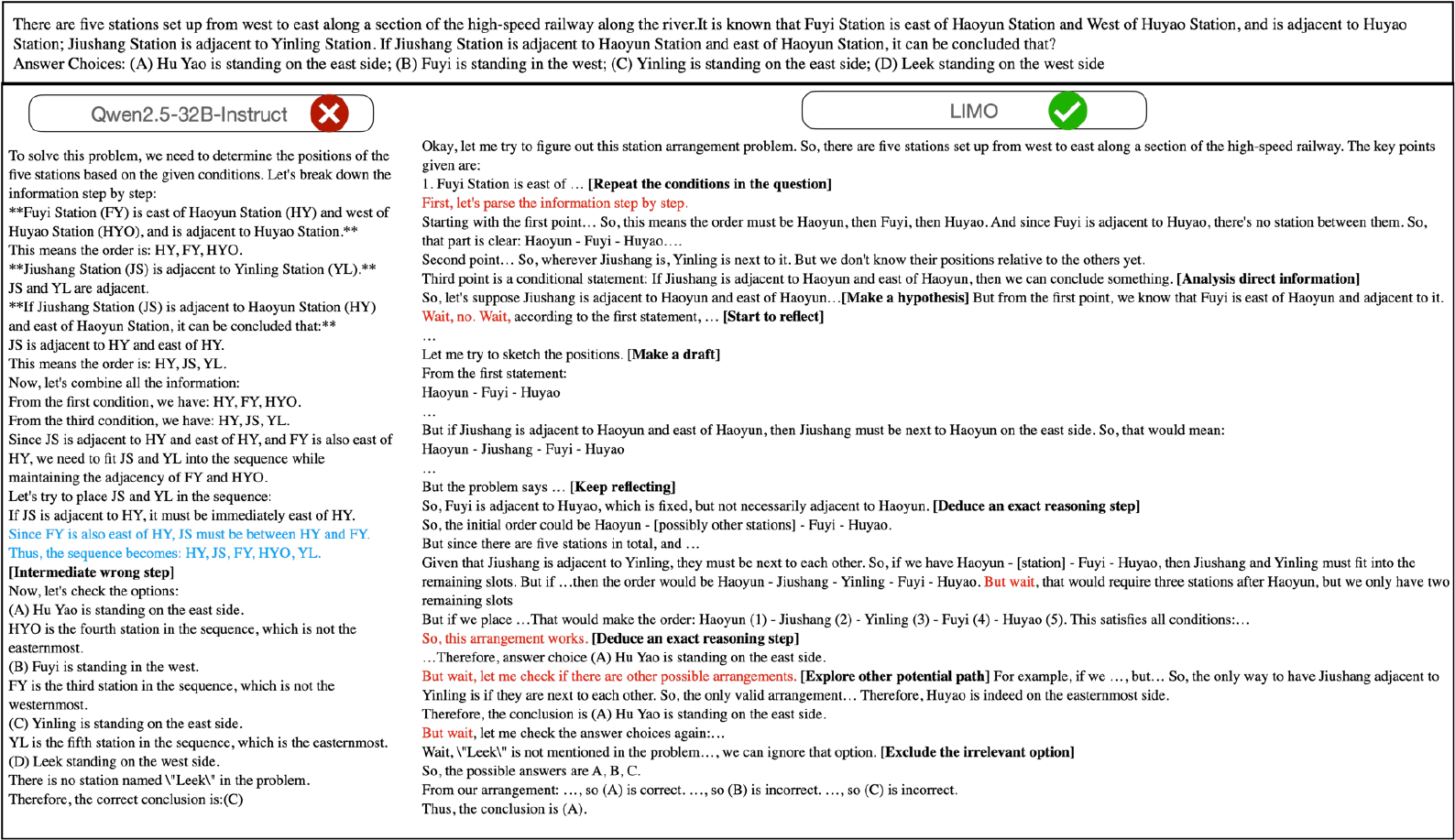}
  \caption{Outputs of Qwen2.5-32B-Instruct and LIMO to a question from LogiQA }
  \label{fig:case_logiqa}
\end{figure*}
Here we present results of the emerging approach nowadays, which distills the models with long and self-reflective reasoning chains.
Previous work has found that training with long reasoning chains on MPS problems only can activate long reasoning patterns for other domains as well~\citep{huang2024o1replicationjourney}, thus it is appealing to study whether this finding holds in our case, and whether the extended length leads to improved performance.
As both LIMO and s1.1 in our evaluation are based on Qwen2.5-32B-Instruct which is very strong on our general reasoning benchmarks and nearly saturates on our MPS tasks, we only keep the most difficult general reasoning tasks in the evaluation and add two another coding-related tasks in the comparison, CRUX~\citep{gu2024crux} that is a benchmark to evaluate code execution output given input and input given the execution output, and HumanEval~\citep{chen2021codex} that aims to assess code generation abilities. For math-related tasks, we only keep the most standard GSM8K and MATH that are not our focus here.

In Table \ref{table:qwen_result} we report both the accuracy and the average response length. In contrast with MPS training with short reasoning data, here MPS training with long reasoning data leads to consistent gains on a broad range of reasoning tasks. 
Also, we observe that the response length is increased significantly on all general reasoning tasks. 
This suggests that longer reasoning paths are beneficial for tasks that involve complex reasoning process, contributing to the model's enhanced performance.
Notably, the LIMO training data is solely composed with mathematical examples and the model outperforms Qwen2.5-32B-Instruct on every general reasoning task by up to 18.1 absolute points.
To further understand how MPS training affects the model's behavior in other domains, we present a case study from LogiQA in Figure~\ref{fig:case_logiqa} for LIMO. LIMO is able to provide a more comprehensive solution by reflecting on intermediate steps and exploring multiple reasoning paths, which allows the model to finally arrive at the correct conclusion, even when faced with some errors in the process. 
On the other hand, the Qwen2.5-32B-Instruct model, which followed a single reasoning path, was more vulnerable to errors in the intermediate steps, any mistake along this path had a direct negative impact on the final answer.

To highlight the importance of long CoT reasoning, we conduct an ablation study using the same question from s1\_1 but with a shorter CoT answer. Specifically, we employ GPT-4o to rewrite the answers from the s1\_1 training data, resulting in an average answer length of approximately 418.9 tokens. The results show that the performance of s1\_1-short is consistently lower than that of the base model, aligning with the conclusion in \textsection\ref{sec:results}.

\subsection{Results of Reinforcement Learning}

\setlength{\tabcolsep}{3pt}
\begin{table*}[!t]
\vspace{-2mm}
\centering
\resizebox{1\linewidth}{!}{
\begin{tabular}{l|cc|cccc}
\toprule
& \multicolumn{2}{c|}{MPS Tasks} & \multicolumn{4}{c}{General Reasoning Tasks} \\
\cmidrule(lr){2-3} \cmidrule(lr){4-7}
\textbf{Model} & GSM8K & MATH & GPQA\_diamond & LogiQA & ARC-c & HumanEval \\
\midrule

Qwen2.5-Math-7B & 91.4(941.6) & 53.5(1290.7) & 44.4(794.4) & 26.6(673.3) & 59.5(214.7) & 64.0(364.5)\\
Qwen2.5-Math-SimpleRL-Zero & 90.6(408.2) & 81.5(535.2) & 48.5(1124.3) & 35.3(1276.9) & 70.7(524.7) & \textbf{67.1}(525.5)\\
Qwen2.5-Math-SimpleRL & \textbf{91.7}(522.1) & \textbf{85.1}(926.8) & \textbf{65.2}(1598.3) & \textbf{45.3}(1302.5) & \textbf{79.7}(611.8) & 50.6(990.8)\\

\midrule
Qwen2.5-7B & 62.9(621.0) & 61.8(655.9) & 39.4(739.2) & 40.9(530.0) & \textbf{80.5}(367.9) & 79.3(199.1)\\
Qwen2.5-7B-SimpleRL & \textbf{91.9}(328.7) & \textbf{78.7}(716.1) & \textbf{51.0}(966.1) & \textbf{41.4}(566.7) & 79.5(347.2) & \textbf{79.8}(434.6)\\

\midrule
Qwen2.5-32B & 82.9(342.4) & 59.2(622.2) & 53.0(469.2) & 54.8(530.7) & 79.7(420.5) & 79.8(494.4)\\
Qwen2.5-32B-SimpleRL & \textbf{96.1}(306.4) & \textbf{85.0}(635.3) & 53.0(819.2) & 	\textbf{63.9}(535.8) & \textbf{92.2}(320.2) & \textbf{91.4}(200.8)\\

\bottomrule
\end{tabular}
}
\caption{Accuracy (tokens in response) of Qwen2.5-Math-7B, Qwen2.5-Math-SimpleRL, Qwen2.5-Math-SimpleRL-Zero, Qwen2.5-7B, Qwen2.5-7B-SimpleRL Qwen2.5-7B and Qwen2.5-7B-SimpleRL.}
\label{table: rl_result}
\vspace{-4mm}
\end{table*}

In this section, we present the results of rule-based RL models trained with MPS samples. 
These RL models are adopted from recent work~\citep{zeng2025simplerl}, employing a similar rule-based reward approach as DeepSeek-R1. Specifically, we include models labeled SimpleRL-Zero, which initiate RL directly from the base model, and SimpleRL models that incorporate an SFT warmup before applying RL, the results are as shown in the Table~\ref{table: rl_result}.

The results demonstrate that RL models indeed exhibit notable generalization capabilities. For instance, SimpleRL and SimpleRL-Zero consistently outperform base model on MPS tasks such as GSM8K and MATH, and also on general reasoning benchmarks like LogiQA and ARC-c. This suggests that reinforcement learning not only enhances the model’s ability to solve mathematical problems but also improves reasoning capabilities beyond the training distribution. 

Besides, SimpleRL outperforms SimpleRL-Zero across most tasks, including a significant improvement on MATH and ARC-c. This finding suggests that incorporating an intermediate SFT stage before RL training is beneficial for enhancing model performance. The SFT stage likely provides a more stable initialization, allowing the RL phase to further refine and reinforce reasoning abilities rather than struggling with unstable policy updates.

\section{Preliminary Study on Searching for Other Tasks for General Reasoning}

So far, we have explored the generalization of various math-related data sources on general reasoning learning. 
In the last part, we conduct a preliminary study to search for other public, commonly used reasoning data sources that are not only about math-related data, to examine whether they can lead to generalized reasoning improvement. 
Specifically, we focus on searching for instruction tuning datasets that are the most widely used to boost models' reasoning performance in practice.

\paragraph{Setup.}

\begin{wraptable}{r}{0.5\textwidth}
\centering
\vspace{-4mm}
\setlength{\tabcolsep}{2pt}
\resizebox{\linewidth}{!}{
\begin{tabular}{lcccc}
\toprule
\textbf{Dataset} & Size & Code & Reasoning & Knowledge \\
\midrule
Magicoder-Evol-Instruct & 110K & \cmark & \xmark & \xmark \\
Magpie-Reasoning        & 150K & \cmark & \cmark & \xmark \\
OpenOrca                & 200K & \cmark & \cmark & \cmark \\
\bottomrule
\end{tabular}
}
\caption{Areas covered by the three selected non-MPS SFT datasets.}
\label{non_mps_data}
\vspace{-2mm}
\end{wraptable}

We identify three non-MPS SFT datasets as our targets to study based on their diverse task coverage and popularity: MagiCoder-Evol-Instruct~\citep{wei2023magicoder}, Magpie-Reasoning~\citep{xu2024magpie}, and OpenOrca~\citep{OpenOrca} We show their area coverage and sizes in Table~\ref{non_mps_data}, and more details can be found in Appendix~\ref{appendix:non-mps-data}. 
Similar to before, we mix these datasets with the UltraChat dataset to perform SFT training. 
We choose \texttt{mistral-7b-v0.1} as our base model.

\paragraph{Results.}
\begin{figure*}[!t]
    \vspace{-2mm}
    \centering
    \includegraphics[width=1\linewidth]{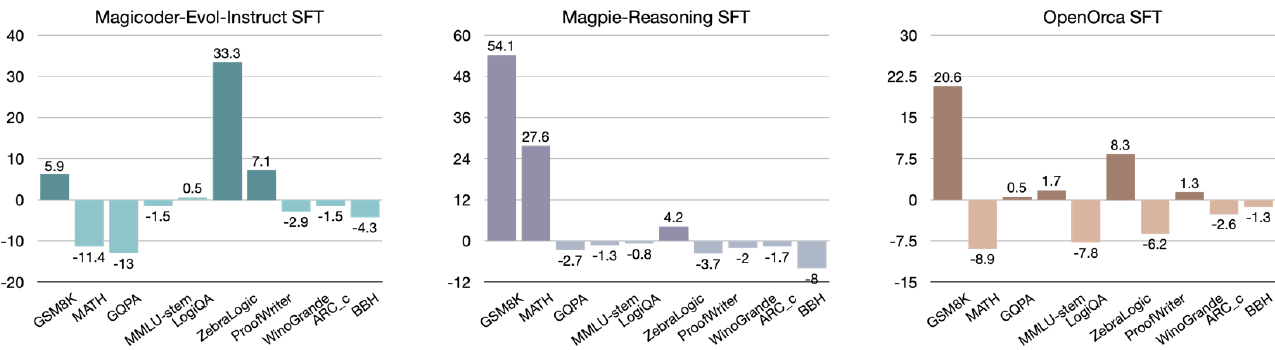}
    \caption{Relative gain (\%) of models trained with instruction tuning on three types of non-MPS datasets compared to the base model trained on UltraChat only across different tasks.}
    \vspace{-2mm}
    \label{fig:fig_non-mps-re}
\end{figure*}
As shown in Figure~\ref{fig:fig_non-mps-re}, despite incorporating diverse datasets, the models trained with non-MPS data still show limited improvements and do not demonstrate broad generalization across different reasoning tasks.  
The performance gains are confined to only a few specific tasks. For example, the Magicoder-Evol-Instruct SFT model shows significant improvement on ZebraLogic, likely due to the dataset focuses on code-related tasks, which aids in addressing logical reasoning problems. The OpenOrca SFT model exhibits more fluctuating gains across different tasks, with some areas showing significant gains and others experiencing declines. This could be due to the complexity and diversity of the OpenOrca dataset, where the model may struggle to balance competing learning objectives across various types of tasks.
Despite these localized improvements, the overall trend indicates that instruction tuning, even with diverse datasets, fails to enhance the model’s generalization abilities across a wide range of reasoning tasks.
These results suggest that new training goals may be required to achieve generalized reasoning. For example, concurrent to our study,~\citet{li2025codeiocondensingreasoningpatterns} demonstrate that predicting code execution output and input is a surprisingly useful training task -- even without long reasoning chains, this training task leads to significant gains on a broad range of out-of-domain reasoning tasks.

\section{Discussion}
In this paper, we explore the generalization potential of different training strategies to learn mathematical problem-solving.  Our experiments reveal that only SFT with long reasoning chains demonstrates consistent generalization to other reasoning tasks, while relatively traditional approaches such as continual pretraining and distillation with short reasoning chains only indicate limited or no generalization gains. Our preliminary study on other, non-MPS training goals fail to achieve satisfactory generalization, implying that learning generalized reasoning is non-trivial with existing open-source datasets.
Future research could explore how different training approaches such as reinforcement learning, which is an appealing direction as recent works demonstrate that RL could generalize better than SFT~\citep{chu2025sftmemorizesrlgeneralizes}. 
Another direction is to search for new training tasks that can potentially generalize better.

\bibliography{colm2025_conference}
\bibliographystyle{colm2025_conference}

\appendix
\section*{Appendix}
\section{Experiment Details}
\label{sec:appendix}

\subsection{Training Hyperparameters}
\label{appendix:training_hyperparameters}
The training process was carried out using the AdamW optimizer with a cosine learning rate scheduler. The training utilized a warmup ratio of 0.1 and set the batch size as 512. Additionally, the training was conducted using DeepSpeed with stage2 configuration. All of these SFT models were fine-tuned using the FastChat~\citep{zheng2023judging} framework with a peak learning rate of 2e-5. 
Based on the FashChat original framework, we also adapted the sequence packaging technique to speed up the training.

For the mix-data training, we trained for 3 epochs for both Math-CoT SFT (mixed) model and Math-PoT SFT (mixed) model. We trained all models on a cluster with 8 NVIDIA A800 GPUs. 

\subsection{Brief Introcution of Benchmarks}
\label{appendix:benchmarks}

Here are the brief introduction to each benchmark. For some complex benchmarks, we also present the corresponding prompt for evaluation.

\paragraph{GSM8K} GSM8K~\citep{cobbe2021gsm8k} is a dataset specifically designed for evaluating LLMs in the domain of multi-step mathematical reasoning. The problem in this dataset are high quality linguistically diverse grade school math word problems created by human problem writers.

\paragraph{GSM8K MQA} This is a dataset where we reformatted the original GSM8K dataset into multiple-choice questions. We kept the original question and let GPT-4o generate other three confusing answers based on the original answer. Models need to generate the option letter of the correct answer.

\paragraph{MATH} MATH~\citep{hendrycksmath2021} test dataset contains 5,000 challenging competition mathematics problems. Each problem in MATH has a full step-by-step solution which can be used to teach models to generate answer derivations and explanations.

\paragraph{MMLU-STEM} We select the original set of STEM tasks from MMLU, excluding direct math problems. Specifically, it include \emph{anatomy, astronomy, college biology, college chemistry, college computer science, college physics, computer security, conceptual physics, electrical engineering, high school biology, high school chemistry, high school computer science, high school physics, high school statistics, machine learning}.

\paragraph{MR-BEN-math} MR-BEN~\citep{zeng2024mrben} is a comprehensive benchmark demands a meta reasoning skill, where LMs are asked to locate and analyse potential errors in automatically generated reasoning steps. We choose the math among all subjects for evaluation.
\begin{tcolorbox}[colback=gray!20, colframe=gray!50, title=MR-BEN-math]

Following is a question and solution pair in subject college math. Your task is to examine the solutions step by step and determine the solution correctness.
If the solution is incorrect, please further find out the first error step and explain the error reason. 
\newline\newline
\textless few-shot examples\textgreater
\newline\newline
Below is the question and solution for you to solve:

Question: \textless question\textgreater

Options: \textless options\textgreater

Please follow the desired response format:

Solution Analysis: [Give a step by step analysis on the solution correctness here]
Solution Correctness: [Input 'correct'/'incorrect' here to indicate the overall correctness of the solution]

First Error Step: [Input 'Step x' here to indicate the first error step here. Input 'N/A' if the solution is correct.]

Error Reason: [Input the error reason and the rectified reasoning of the first error step here. Input 'N/A' if the solution is correct.]
\newline\newline
Please follow this format without any additional introductory or concluding statements.
\end{tcolorbox}

\paragraph{DocMath} DocMath~\citep{zhao2024docmatheval} is a benchmark specifically designed to evaluate the numerical reasoning capabilities of LLMs in the context of understanding and analyzing specialized documents containing both text and tables. Models are asked to generate answer through CoT.
\begin{tcolorbox}[colback=gray!20, colframe=gray!50, title=DocMath]
You are a financial expert, you are supposed to answer the given question based on the provided financial document context. You need to first think through the problem step by step, documenting each necessary step. Then you are required to conclude your response with the final answer in your last sentence as 'Therefore, the answer is {final answer}'. The final answer should be a numeric value.
\newline

USER: \textless context and document\textgreater

Question: \textless question \textgreater

Let's think step by step to answer the given question.

ASSISTANT:
\end{tcolorbox}

\paragraph{ZebraLogic} ZebraLogic~\citep{zebralogic2024} is a benchmark consisting of logic grid puzzles and evaluates the logical reasoning abilities of LLMs. Each puzzle presents N houses with M features, requiring unique value assignments based on given clues. We use the average result of LLMs of different levels of puzzles.
\begin{tcolorbox}[colback=gray!20, colframe=gray!50, title=ZebraLogic]
A chat between a curious user and an artificial intelligence assistant. The assistant gives helpful, detailed, and polite answers to the user's questions. 
\newline

USER: 

\# Puzzle to Solve

\textless puzzle\textgreater

\#\# Clues:

\textless clues\textgreater

\# Instruction

Now please solve the above puzzle. Present your reasoning and solution in the following json format:

\textless output format\textgreater
\end{tcolorbox}

\paragraph{LogiQA} LogiQA~\citep{liu2020logiqa} is a benchmark which is sourced from expert-written questions for testing human Logical reasoning, covering multiple types of deductive reasoning.

\paragraph{ProofWriter} Proofwriter~\citep{tafjord2020proofwriter} contains many small rule-bases of facts and rules, expressed in English. Each rule-base also has a set of questions which can either be proven true or false using proofs of various depths, or the answer is ``Unknown'' or assumed negative. 
\begin{tcolorbox}[colback=gray!20, colframe=gray!50, title=ProofWriter]
Task Description: You are given a problem description and a question. The task is to: 

1) define all the predicates in the problem

2) parse the problem into logic rules based on the defined predicates

3) write all the facts mentioned in the problem

4) parse the question into the logic form

------

\textless few-shot examples\textgreater

------
\newline

Problem:\newline
[[PROBLEM]]\newline
Question:\newline
[[QUESTION]\newline
\#\#\#

\end{tcolorbox}

\paragraph{GPQA} GPQA~\citep{rein2023gpqa} is a multiple-choice, Q\&A dataset of very hard questions written and validated by experts in biology, physics, and chemistry.

\paragraph{WinoGrande} WinoGrande~\citep{sakaguchi2021winogrande} is designed for commonsense reasoning. The samples are formulated as fill-in-the-blank questions where two answer choices are provided. The goal is to select the correct option based on commonsense knowledge.

\paragraph{ARC-challenge} AI2 Reasoning Challenge (ARC)~\citep{allenai:arc} is a widely used dataset for evaluating large language models (LLMs) on their commonsense reasoning abilities. We choose the challenge set of ARC, which contains questions that simple retrieval or co-occurrence-based models struggle with, thus pushing models to reason more deeply.

\paragraph{BBH} BBH~\citep{suzgun2022challenging} is a challenging benchmark designed to evaluate the reasoning capabilities of language models across a wide range of tasks. It includes tasks that test general reasoning, such as commonsense reasoning, logical reasoning, symbolic reasoning and so on, with an emphasis on harder, more complex problems.

\paragraph{CRUX} CRUXEval is a benchmark designed to evaluate code language models (LMs), consisting of 800 Python functions along with corresponding input-output pairs. The benchmark includes two tasks: CRUXEval-I (input prediction) and CRUXEval-O (output prediction), both of which assess the model's ability to reason, understand, and execute code.

\paragraph{HumanEval} HumanEval is a dataset widely used for evaluating code generation models, particularly in programming tasks. The dataset is designed to assess a model's ability to reason through programming problems and execute code to verify correctness. The dataset includes a series of programming problems that require models to understand the problem, reason about the solution, and generate code that can be executed to check for correctness.

\subsection{Non-MPS Instruction Tuning Datasets}
\label{appendix:non-mps-data}

Considered that short math CoT data have demonstrated limited success in improving the model's ability to generalize, especially for non-mathematical reasoning tasks, we explore three non-MPS datasets with the aim of enhancing the model’s performance across a broader spectrum of reasoning tasks.

In particular, we focus on three non-MPS SFT datasets, chosen for their wide-ranging task coverage. Details are as followed:

\textbf{Magicoder-Evol-Instruct\footnote{\href{https://huggingface.co/datasets/ise-uiuc/Magicoder-Evol-Instruct-110K}{https://huggingface.co/datasets/ise-uiuc/Magicoder-Evol-Instruct-110K}}}~\citep{wei2023magicoder} is used primarily to enhance code generation capabilities in LLMs. The dataset was decontaminated and repurposed from an earlier open-source instruction dataset, Evol-CodeAlpaca\footnote{\href{https://huggingface.co/datasets/theblackcat102/evol-codealpaca-v1}{https://huggingface.co/datasets/theblackcat102/evol-codealpaca-v1}}, which has augmented questions and answers by GPT-4.The dataset helping improve the performance of LLMs on code generation and program algorithm tasks, particularly in diverse programming contexts. 

\textbf{Magpie-Reasoning\footnote{\href{https://huggingface.co/datasets/Magpie-Align/Magpie-Reasoning-150K}{https://huggingface.co/datasets/Magpie-Align/Magpie-Reasoning-150K}}} is a specialized SFT dataset designed to improve the reasoning capabilities of LLMs. It is generated by Qwen2-72B-Instruct~\citep{yang2024qwen2} and Llama-3-70B Instruct~\citep{dubey2024llama3herdmodels} using Magpie~\citep{xu2024magpie}. It consists of 150K samples of conversations, covering a mix of tasks including mathematical reasoning, code-based reasoning, and general logic-based problem-solving.

\textbf{OpenOrca\footnote{\href{https://huggingface.co/datasets/Open-Orca/OpenOrca}{https://huggingface.co/datasets/Open-Orca/OpenOrca}}} is a large, open-domain dataset that spans diverse fields, including math, science, general knowledge, and other multi-domain tasks, with the distributions outlined in Orca~\citep{mukherjee2023orca}. This dataset is augmented from FLAN collection data~\citep{longpre2023flan} with GPT-4.  Given resource limitations, we performed SFT on 200K samples.

\end{document}